\documentclass{article}
\usepackage{spconf,amsmath,graphicx}

\usepackage{array}
\usepackage{rotating}
\usepackage{makecell}
\usepackage{longtable}
\usepackage{tabularx}
\usepackage{tabu}
\usepackage{bigstrut}
\usepackage{rotating}
\usepackage{multirow}
\usepackage{multicol}
\usepackage{float, color}
\usepackage[sort&compress,numbers,comma]{natbib}

\title{Spatial-context-aware deep neural network for multi-class image classification}

\makeatletter
\def\@name{Jialu ZHANG$^{* \dagger}$, Qian ZHANG$^{*}$, Jianfeng REN$^{*}$, Yitian ZHAO$^{\dagger}$, Jiang LIU$^{*\dagger \ddagger}$\thanks{This work was supported in part by the National Natural Science Foundation of China under Grant 72071116, and in part by the Ningbo Municipal Bureau of Science and Technology under Grant 2019B10026.}}

\address{$^{*}$ School of Computer Science, University of Nottingham Ningbo China \\$^{\dagger}$ Cixi Institute of Biomedical Engineering, Chinese Academy of Sciences \\$^{\ddagger}$ Department of Computer Science and Engineering, Southern University of Science and Technology}

\begin{document}
%
\maketitle
\begin{abstract}
	Multi-label image classification is a fundamental but challenging task in computer vision. Over the past few decades, solutions exploring relationships between semantic labels have made great progress. However, the underlying spatial-contextual information of labels is under-exploited. To tackle this problem, a spatial-context-aware deep neural network is proposed to predict labels taking into account both semantic and spatial information. This proposed framework is evaluated on Microsoft COCO and PASCAL VOC, two widely used benchmark datasets for image multi-labelling. The results show that the proposed approach is superior to the state-of-the-art solutions on dealing with the multi-label image classification problem.
\end{abstract}
%
\begin{keywords}
 Multi-label, image classification, deep learning, spatial context
\end{keywords}

\section{Introduction}
\label{sec:intro}
\vspace{-0.2cm}
With the rapid development of technologies, abundant visual information is constantly received. One of the fundamental but challenging problems for image understanding is to label the objects, locations or attributes in the images, possibly with more than one label. Multi-label image classification problem has attracted the attention of researchers in the past few years. However, the rich semantic information and higher-order label co-occurrence are challenging to model \cite{liu2017survey, silla2011survey}. 

Recently, many deep convolutional neural networks (DCNNs) were developed for single-label image classification problem \cite{simonyan2014very,szegedy2015going,he2016deep}, and transforming the multi-label classification problem into multiple binary classification tasks is one of the common strategies \cite{tsoumakas2007multi} to solve the multi-label image classification problem. However, this kind of method ignores the inter-dependencies among labels, which have been proved useful \cite{wang2016cnn, yeh2017learning}. To tackle this problem, researchers developed various DCNNs \cite{wang2016cnn, 8682549, zhu2017learning, Guo_2019_CVPR, 8683665, Chen_2019_CVPR, Lanchantin_2021_CVPR} that can consider all the labels for a given image concurrently. For example, Wang \emph{et al.} designed a sequentially predict model and used a recurrent neural network (RNN) to determine label dependencies \cite{wang2016cnn}, while Chen \emph{et al.} and Wang \emph{et al.} employed graph convolutional networks to capture global label dependencies \cite{Chen_2019_CVPR, wang2019multilabel}. By exploiting label-correlation information, these approaches made great progress on image multi-labelling. Nevertheless, object spatial information and image context information are not fully exploited in these approaches.

\begin{figure}[t]
	\centering
	\centerline{\includegraphics[width=0.6\linewidth]{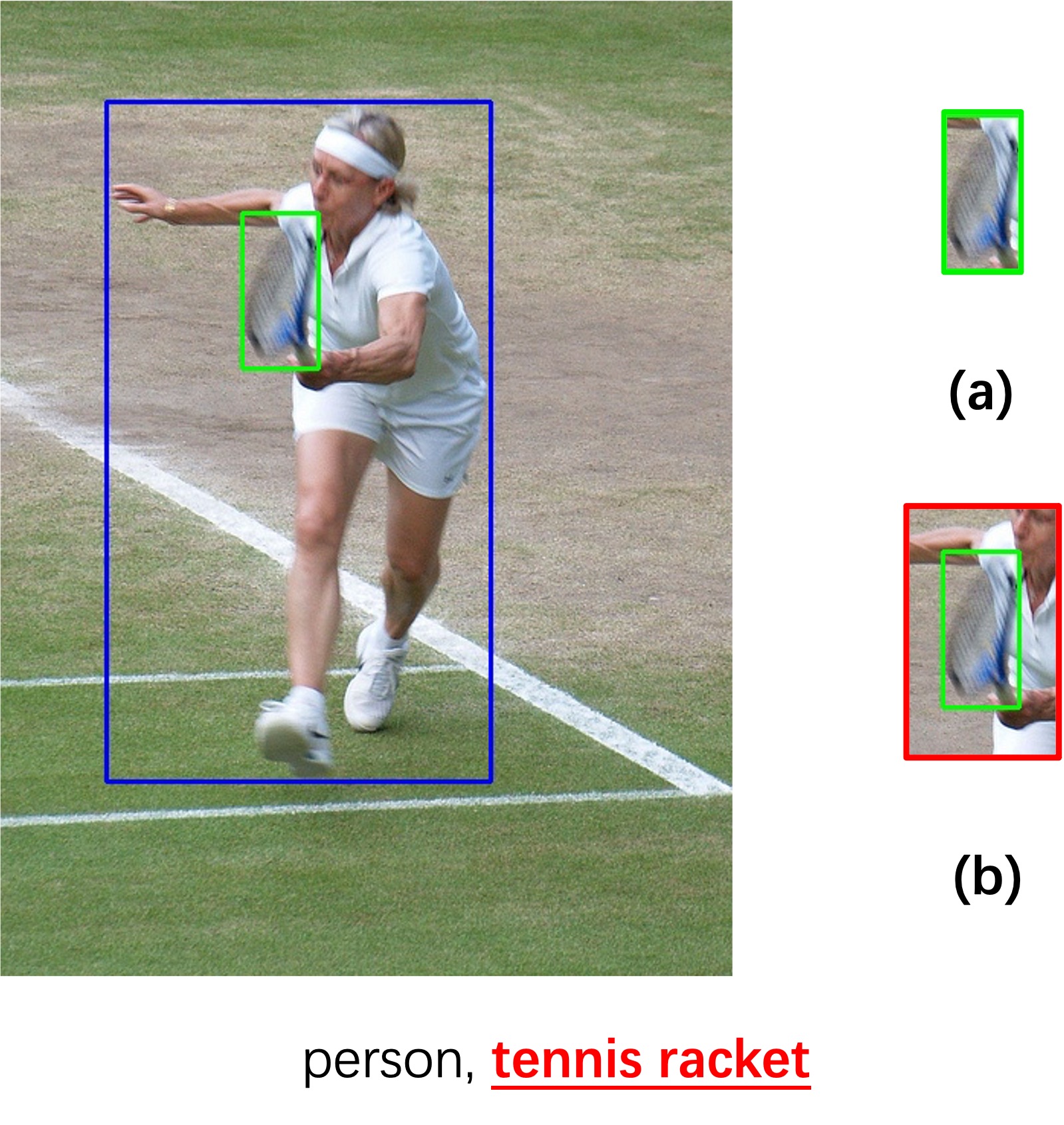}}
 \vspace{-0.4cm}
 \caption{An illustrative example. Previous methods cannot detect the tennis racket. By exploiting the label dependencies, spatial and context information, the proposed approach could more accurately detect it.}
 \label{fig:example}
 \vspace{-0.6cm}
\end{figure}

Wei \emph{et al.} \cite{wei2014cnn} and Yang \emph{et al.} \cite{yang2016exploit} addressed this problem by devising a 2-stage pipeline for multi-labelling in which the model generates image patches first and then labels them. However, these methods overemphasize the generated patches, thereby neglecting surrounding context infomation and label dependencies. The idea of object localization is similar to the attention mechanism that has been successfully applied in many vision tasks \cite{zhu2017learning, Guo_2019_CVPR, wen2020multilabel, you2020crossmodality, 8682335}. Fig. \ref{fig:example} illustrates the importance of label dependencies, spatial and context information. Additionally, context has been demonstrated useful in various visual processing tasks, such as recognition and detection \cite{yang2021rain, ren2015learning, MOHAMMADI2020107303}.

To make good use of these dependencies and information, a two-branch spatial-context-aware DCNN is designed, where one branch is designed to extract the spatial information of the objects and the other aims at capturing the image context information. The network’s label predictor follows the principle of multi-label image classification and utilizes the dependencies among labels. Moreover, with more contextual information, the proposed model performs well in labelling small objects that other models may not be able to capture. 

Different from existing spatial-context-aware models exploiting the context information between objects \cite{bardool2019systematic} or considering the receptive field of the shallow layers as the context information to the pixels on the deep layers of the network \cite{KONG2021107867}, the proposed method exploits the spatial context information directly on the feature maps and utilizes the feature maps around the object as the background context. The experimental results on two large benchmark datasets, MS-COCO and PASCAL VOC, demonstrate the effectiveness of the proposed approach compared with state-of-the-art approaches. 

The contributions of this paper are summarized as follow: 1) To make use of the spatial and context information to the object, a two-branch spatial-context-aware deep neural network is proposed for multi-label image classification problem. 2) The proposed image-context-aware branch could well exploit both spatial and semantic information of objects. 3) The proposed approach significantly outperforms the state-of-the-art approaches \cite{wang2016cnn, he2016deep, zhu2017learning, Guo_2019_CVPR, Chen_2019_CVPR, wen2020multilabel, wang2019multilabel, you2020crossmodality, liu2021weakly, Lanchantin_2021_CVPR} on the MS-COCO dataset \cite{lin2015microsoft} and PASCAL VOC \cite{pascal-voc-2007} dataset.

\section{Methodology}
\label{sec:methodology}

\subsection{Overall Framework}
\begin{figure}[h]
 \centering
 \centerline{\includegraphics[width=8.8cm]{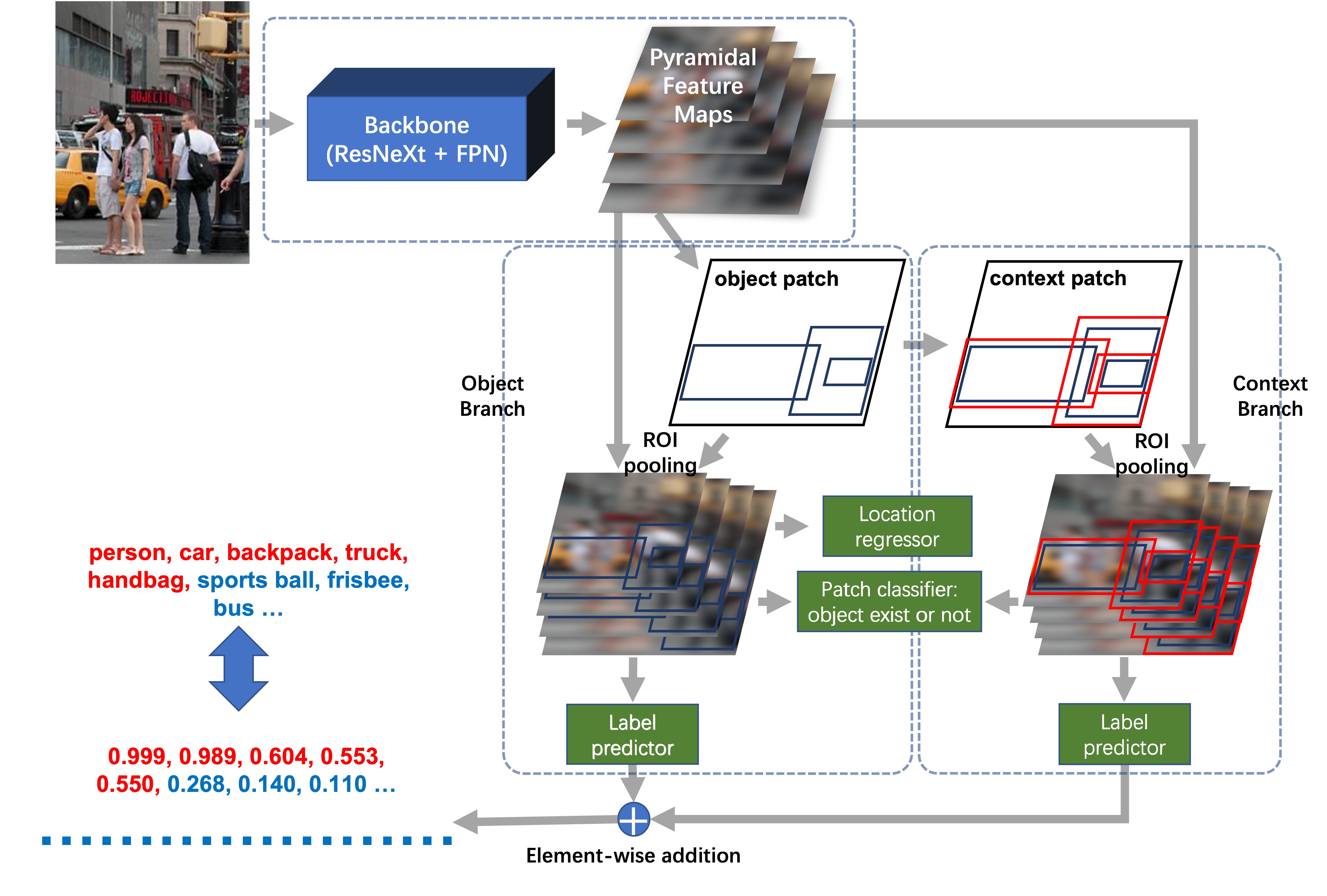}}
 \caption{An overview of the proposed model.}
 \label{fig:structure}
\end{figure}

The overall framework of the proposed model is presented in Fig. \ref{fig:structure}. It consists of three main modules: 1) A feature extractor which integrates ResNeXt-101 \cite{xie2017aggregated} with feature pyramid networks (FPN) \cite{lin2017feature}. 2) Two patch generators. To utilize both spatial and contextual information, the object branch follows the structure in \cite{ren2016faster} to localize the objects and the context branch considers the image context information. These fragments and feature maps are combined to generate the final visual features. 3) Four patch processors. One regressor for positioning, one classifier for determining the existence of objects in the patch, and two predictors for predicting the label of the patch using a formulation of image multi-labelling.

The patch labelling results from the object branch and context branch are combined by an element-wise classifier fusion to generate the final predicted labels. 

\subsection{Feature Extractor}
ResNeXt-101 \cite{xie2017aggregated} followed by FPN \cite{lin2017feature} is utilized as the feature extractor. The former aggregates a set of transformations to improve the classification capabilities of deep neural networks, and the latter employs the pyramid representations to extract a rich visual semantic abstract. The last pooling and classification layers in ResNeXt-101 are removed and the feature maps from the last convolutional layer are used as the input of FPN. A 4-stage semantic feature pyramid is built in FPN from high to low resolution. Denote the two successive feature extracting networks as $f_{R}\left(\mathbf{U} ; \theta_{R}\right)$ and $f_{F}\left(\mathbf{Z} ; \theta_{F}\right)$ respectively. Given an input image $\mathbf{I}$, the final generated pyramidal feature $\mathbf{X}$ can be obtained as:

\begin{equation}
 \mathbf{X}= f_{F}\left(f_{R}\left(\mathbf{I} ; \theta_{R}\right) ; \theta_{F}\right).
\end{equation}

\subsection{Patch Generators}

Based on the generated feature maps, a set of bounding boxes are generated to locate objects, covering both the contextual and spatial information of the objects. However, a tightly cropped bounding box may only contain the information of the object, but ignore the contextual information surrounding the object.

The tightly cropped bounding box in the object branch may neglect the contextual information, so the context branch is utilized to expand it into a larger one with potential object and its surrounding image content. Note that the expanded context patch is only used to determine the label of the object, but not for locating the object. As Fig. \ref{fig:example} shows, confidence in predicting `tennis racket' increases as information about the surrounding environment, e.g., `person', is included.

The tightly cropped object patch contains the most discriminant information for classify the object, while the expanded image patch containing both the object and additional contextual information. Hence, two branches with different objectives are designed to generate image patches. Specifically, as shown in Fig. \ref{fig:structure}, the generator in the object branch (left) takes the feature maps as the input and produces a set of rectangular object patches with confidence scores. Only objects within the rectangles are evaluated in this branch, and it is often difficult to detect small objects. Therefore, a patch expansion mechanism is applied to the generator in the context branch. This design allows more contextual information to be captured, and hence improves the accuracy of the patch prediction. Two branches are used together to predict the labels of the input image.

\subsection{Multi-task Patch Processors}

The generated object patches and context patches are then fed into four dense networks for location regression, patch classification and label prediction, respectively.

The first dense network, location regressor, is designed to accurately locate the objects to explore and utilize the image-level spatial information of different labels. It is guided by the location regression loss $L_{r}$. The objective is to maximize the intersection over union (IOU) between the generated bounding boxes and the ground-truth bounding boxes.

\begin{equation}
 L_{r}\left(\hat{t}_{i}, t_{i}^{*}\right)=\sum_{i \in\{\mathrm{x}, \mathrm{y}, \mathrm{w}, \mathrm{h}\}}\phi\left(\hat{t}_{i}-t_{i}^{*}\right),
\end{equation}

\begin{equation}
 \phi(x)=\left\{\begin{array}{ll}
 0.5 x^{2} & \text { if }|x|<1 \\
 |x|-0.5 & \text { otherwise }
 \end{array}\right.,
\end{equation} where $\hat{t}_{i}$ and $t_{i}^{*}$ represent the predicted patches and ground-truth bounding boxes, respectively and x, y, w, h correspond to the centre coordinates, height and width of the proposed patch. $\phi\left(x\right)$ is a non-sensitive outlier-removal function for enhancing the training robustness.


The second dense network, patch classifier, determines the confidence whether an object exists in the bounding box. Hence it is a binary classification problem. The cross-entropy loss is used to guide the training process and the loss is defined as follows:

\begin{equation}
 L_{p}\left(\hat{p}, p^{*}\right) = -p^{*} \log \hat{p}+\left(1-p^{*} \right) \log \left(1-\hat{p}\right),
\end{equation} where $\hat{p}$ and $p^{*}$ represent the output confidence from the network and ground-truth, respectively. $p^{*}$ would be assigned to $1$ if the maximum IOU between the current patch and the ground-truth one exceeds a certain threshold, and 0 otherwise.


The remaining two dense networks are label predictors to determine which object the bounding box contains. These two dense networks are trained by using the binary cross-entropy (BCE) loss, which can exploit label dependencies in the multi-labelling tasks \cite{zhu2017learning, wang2016cnn, yeh2017learning}. 

\sloppy Given a one-hot vector of ground-truth labels $\mathbf{y^{*}} = \left[y^{1}, y^{2}, \ldots, y^{C}\right]^{T}$, where $C$ represents the number of all possible labels in the dataset. $y^{i}$ is a binary indicator that $y^{l} = 1$ if the image contains the label $l$, and $y^{l} = 0$ otherwise. The label-prediction loss is denoted as follows:

\begin{equation}
 \begin{split}
 L_{l}\left(\mathbf{\hat{y}}, \mathbf{y^{*}}\right)=\sum_{i=1}^{C} y^{i} \log \sigma\hat{y}^{i}+\left(1-y^{i}\right) \log \left(1-\sigma\hat{y}^{i}\right).
 \end{split}
\end{equation} Similar to $\mathbf{y^{*}}$ and $y^{i}$, $\mathbf{\hat{y}}$ and $\hat{y}^{i}$ denote the predicted confidence over all possible labels and the $i$-th category. $\sigma$ in $L_{l}$ is a learnable weighting factor, we fix it as 0.5 given by the empirically study.

Hence the total loss function $L$ is represented as follows:

\begin{equation}
 L= L_{r}+ \alpha L_{p}+ \beta L_{l}^{O} + \gamma L_{l}^{C},
\end{equation} where $\alpha, \beta$ and $\gamma$ is scale values that utilized to balance the losses and $L_{l}^{O}$ and $L_{l}^{C}$ represent the loss of the object branch and context branch respectively.

\section{Experimental Results}
\label{sec:experiments}

The proposed method is compared with various state-of-the-art models on two benchmark datasets - Microsoft COCO-2017 \cite{lin2015microsoft} and PASCAL VOC-2007 \cite{pascal-voc-2007}. Experimental results indicate that the proposed model significantly and consistently outperforms all the compared solutions.
 
\begin{table*}[t]
 \centering
 \resizebox{0.81\textwidth}{!}{
 \begin{tabular}{p{17em}|c|c|c|c|c|c|c}
 \hline
 \multicolumn{1}{p{17em}<{\centering}|}{Method} & \multicolumn{1}{p{4.5em}<{\centering}|}{mAP} & \multicolumn{1}{p{4.5em}<{\centering}|}{F1-C} & \multicolumn{1}{p{4.5em}<{\centering}|}{P-C} & \multicolumn{1}{p{4.5em}<{\centering}|}{R-C} & \multicolumn{1}{p{4.5em}<{\centering}|}{F1-O} & \multicolumn{1}{p{4.5em}<{\centering}|}{P-O} & \multicolumn{1}{p{4.5em}<{\centering}}{R-O} \bigstrut\\
 \hline
 \hspace{0.4em}CNN-RNN (CVPR, 2016, \cite{wang2016cnn}) & 61.2 & 60.4 & 66.0 & 55.6 & 67.8 & 69.2 & 66.4 \bigstrut[t]\\
 \hspace{0.4em}ResNet101 (CVPR, 2016, \cite{he2016deep}) & 75.2 & 69.5 & 80.8 & 63.4 & 74.4 & 82.2 & 68.0 \\
 \hspace{0.4em}RNN-Attention (ICCV, 2017, \cite{Wang_2017_ICCV}) & - & 67.4 & 79.1 & 58.7 & 72.0 & 84.0 & 63.0 \\
 \hspace{0.4em}ResNet101-SRN (CVPR, 2017, \cite{zhu2017learning}) & 77.1 & 71.2 & 81.6 & 65.4 & 75.8 & 82.7 & 69.9 \\
 \hspace{0.4em}RNN-frequency (TMM, 2019, \cite{8624407}) & 64.7 & - & - & - & - & - & - \\
 \hspace{0.4em}DELTA (PR, 2019, \cite{yu2019delta}) & 71.3 & - & - & - & - & - & - \\
 \hspace{0.4em}ResNet101-ACfs (CVPR, 2019, \cite{Guo_2019_CVPR}) & 77.5 & 72.2 & 77.4 & 68.3 & 76.3 & 79.8 & 73.1 \\
 \hspace{0.4em}DecoupleNet (ICASSP, 2019, \cite{8683665}) & 82.2 & 76.3 & 83.1 & 71.6 & 79.5 & 84.7 & 74.8 \\
 \hspace{0.4em}ML-GCN (CVPR, 2019, \cite{Chen_2019_CVPR}) & 83.0 & 78.0 & 85.1 & 72.0 & 80.3 & 85.8 & 75.4 \\
 \hspace{0.4em}ResNet101-CRL (TSMC-S, 2020, \cite{wen2020multilabel}) & 81.1 & 75.8 & 81.2 & 70.8 & 78.1 & 83.6 & 73.3 \\
 \hspace{0.4em}KSSNet (AAAI, 2020, \cite{wang2019multilabel}) & 83.7 & 77.2 & 84.6 & 73.2 & 81.5 & \textbf{87.8} & 76.2 \\
 \hspace{0.4em}MS-CMA (AAAI, 2020, \cite{you2020crossmodality}) & 83.8 & 78.4 & 82.9 & 74.4 & 81.0 & 84.4 & 77.9 \\
 \hspace{0.4em}WSL-GCN (PR, 2021, \cite{liu2021weakly}) & 84.8 & - & - & - & - & - & - \\
 \hspace{0.4em}C-Tran (CVPR, 2021, \cite{Lanchantin_2021_CVPR}) & 85.1 & 79.9 & \textbf{86.3} & 74.3 & 81.7 & 87.7 & 76.5 \bigstrut[b]\\
 \hline
 \hspace{0.4em}The Proposed & \textbf{86.0} & \textbf{80.3} & 84.0 & \textbf{77.5} & \textbf{83.2} & 85.9 & \textbf{80.6} \bigstrut\\
 \hline
 \end{tabular}%
 } 
 \caption{Comparison results on the MS-COCO dataset. The best results are shown in bold.}
 \label{tab:results}%
\end{table*} 

\begin{table*}[htbp]
 \centering
 \resizebox{\textwidth}{!}{
 \begin{tabular}{l|ccccccccccccccccccccc}
 \hline
 Method & areoplane & bicycle & bird & boat & bottle & bus & car & cat & chair & cow & diningtable & dog & horse & motorbike & person & pottedplant & sheep & sofa & train & tvmonitor & \textbf{mAP} \bigstrut[t]\\
 \hline
 CNN-RNN (CVPR, 2016, \cite{wang2016cnn}) & 96.7 & 83.1 & 94.2 & 92.8 & 61.2 & 82.1 & 89.1 & 94.2 & 64.2 & 83.6 & 70.0 & 92.4 & 91.7 & 84.2 & 93.7 & 59.8 & 93.2 & 75.3 & \textbf{99.7 } & 78.6 & 84.0 \bigstrut[t]\\
 ResNet101 (CVPR, 2016, \cite{he2016deep}) & 99.5 & 97.7 & 97.8 & 96.4 & 65.7 & 91.8 & 96.1 & 97.6 & 74.2 & 80.9 & 85.0 & 98.4 & 96.5 & 95.9 & 98.4 & 70.1 & 88.3 & 80.2 & 98.9 & 89.2 & 89.9 \\
 RNN-Attention (ICCV, 2017, \cite{Wang_2017_ICCV}) & 98.6 & 97.4 & 96.3 & 96.2 & 75.2 & 92.4 & 96.5 & 97.1 & 76.5 & 92.0 & 87.7 & 96.8 & 97.5 & 93.8 & 98.5 & 81.6 & 93.7 & 82.8 & 98.6 & 89.3 & 91.9 \\
 RNN-frequency (TMM, 2019, \cite{8624407}) & 97.0 & 92.5 & 93.8 & 93.3 & 59.3 & 82.6 & 90.6 & 92.0 & 73.4 & 82.4 & 76.6 & 92.4 & 94.2 & 91.4 & 95.3 & 67.9 & 88.6 & 70.1 & 96.8 & 81.5 & 85.6 \\
 DELTA (PR, 2019, \cite{yu2019delta}) & 98.2 & 95.1 & 95.8 & 95.7 & 71.6 & 91.2 & 94.5 & 95.9 & 79.4 & 92.5 & 85.6 & 96.7 & 96.8 & 93.7 & 97.8 & 77.7 & 95.0 & 81.9 & 99.0 & 87.9 & 91.1 \\
 ML-GCN (CVPR, 2019, \cite{Chen_2019_CVPR}) & 99.5 & 98.5 & 98.6 & 98.1 & 80.8 & 94.6 & 97.2 & 98.2 & 82.3 & 95.7 & 86.4 & 98.2 & 98.4 & 96.7 & 99.0 & 84.7 & 96.7 & 84.3 & 98.9 & 93.7 & 94.0 \\
 ResNet101-CRL (TSMC-S, 2020, \cite{wen2020multilabel}) & \textbf{99.9 } & 98.4 & 97.8 & \textbf{98.8 } & 81.2 & 93.7 & 97.1 & 98.4 & \textbf{82.7 } & 94.6 & 87.1 & 98.1 & 97.6 & 96.2 & 98.8 & 83.2 & 96.2 & 84.7 & 99.1 & 93.5 & 93.8 \\
 WSL-GCN (PR, 2021, \cite{liu2021weakly}) & 99.7 & 98.5 & \textbf{99.0 } & 97.8 & 86.2 & 96.2 & 98.3 & \textbf{99.3 } & 81.1 & 95.9 & 88.0 & \textbf{99.2 } & \textbf{98.6 } & 97.1 & \textbf{99.4 } & 85.0 & \textbf{97.5 } & 84.3 & 99.0 & 94.0 & 94.7 \bigstrut[b]\\
 \hline
 The Proposed & 99.4 & \textbf{98.8 } & 98.0 & 98.6 & \textbf{90.5 } & \textbf{98.3 } & \textbf{98.6 } & 98.4 & 81.3 & \textbf{96.2 } & \textbf{88.6 } & 96.7 & \textbf{98.6 } & \textbf{99.0 } & 99.3 & \textbf{87.0 } & \textbf{97.5 } & \textbf{87.3 } & 98.6 & \textbf{95.7 } & \textbf{95.3 } \bigstrut\\
 \hline
 \end{tabular}%
 } 
 \caption{Comparison results on the PASCAL VOC dataset. The best results are shown in bold.}
 \label{tab:results_pascal}%
\vspace{-0.2cm}
\end{table*}

\subsection{Experimental Settings}

The proposed model is trained on PyTorch \cite{paszke2019pytorch}. Stochastic gradient descend strategy is employed for training, with a mini-batch size of 2, a momentum of 0.5 and a decay rate of 0.01. The initial learning rate is 0.002. All images are resized to 416$\times$416 before passing to the model. The proposed model is optimized by using the transfer-learning techniques, i.e., load and freeze the pre-trained weights of the backbone, and then train the remaining modules. Once the training converges, the whole model is jointly optimized.

The MS-COCO dataset \cite{lin2015microsoft} and PASCAL VOC \cite{pascal-voc-2007} are two widely used public benchmark datasets for image labelling. The former contains 80 object categories and the latter covers 20 different categories. All instances in the two datasets are associated with at least one label. The standard evaluation protocol is used as in \cite{Guo_2019_CVPR, Chen_2019_CVPR, wen2020multilabel, wang2019multilabel, you2020crossmodality, liu2021weakly, Lanchantin_2021_CVPR}. 


Standard evaluation metrics such as mean average precision (mAP), F1-score, precision and recall are used in experiments. The prefix `macro-' indicates the average over all categories, while `micro-' indicates the average over all samples. Therefore, `macro-' is susceptible to the rare categories and `micro-' is easily dominated by the major classes \cite{tang2009large}. Denote `macro-' as `-C' and `micro-' as `-O'.

The proposed method is compared against state-of-the-art models including CNN-RNN \cite{wang2016cnn}, ResNet101 \cite{he2016deep}, RNN-Attention \cite{Wang_2017_ICCV}, ResNet101-SRN \cite{zhu2017learning}, RNN-frequency \cite{8624407}, DELTA \cite{yu2019delta}, ResNet101-ACfs \cite{Guo_2019_CVPR}, DecoupleNet \cite{8683665}, ML-GCN \cite{Chen_2019_CVPR}, ResNet101-CRL \cite{wen2020multilabel}, KSSNet \cite{wang2019multilabel}, MS-CMA \cite{you2020crossmodality}, WSL-GCN \cite{liu2021weakly} and C-Tran \cite{Lanchantin_2021_CVPR}, among which CNN-RNN \cite{wang2016cnn}, DecoupleNet \cite{8683665}, ML-GCN \cite{Chen_2019_CVPR}, KSSNet \cite{wang2019multilabel} and WSL-GCN \cite{liu2021weakly} exploit semantic relations by capturing global label dependencies. ResNet101 \cite{he2016deep} is initially designed for single-label image classification while also performs well in multi-labelling by using appropriate loss functions. RNN-Attention \cite{Wang_2017_ICCV}, ResNet101-SRN \cite{zhu2017learning}, RNN-frequency \cite{8624407}, DELTA \cite{yu2019delta}, ResNet101-ACfs \cite{Guo_2019_CVPR} and ResNet101-ACfs \cite{Guo_2019_CVPR} exploit spatial information by extending attention ideas, while ResNet101-CRL constructs explicit context relations by feature-label co-projection. MS-CMA \cite{you2020crossmodality} adapts the attention mechanism to a cross-modality version with graph embedding. C-Tran \cite{Lanchantin_2021_CVPR} exploits the dependencies among visual features and labels to tackle the image multi-labeling.

\vspace{-0.1cm}
\subsection{Comparisons to State-of-the-art Approaches}
\vspace{-0.1cm}

The comparison results to the state-of-the-art approaches on the MS-COCO dataset are summarized in Table \ref{tab:results}. It is clear to see that the proposed model significantly outperforms all the state-of-the-art models in terms of the key evaluation metrics such as mAP and F1-score. 

As shown in bold in Table 1, the proposed method achieves an overall mAP of 86.3$\%$, a per-class F1 score of 80.6$\%$ and an overall F1-score of 83.1$\%$, which greatly surpasses the previous best solution, C-Tran \cite{Lanchantin_2021_CVPR} by 1.2$\%$, 0.7$\%$ and 1.4$\%$, respectively. It shows that the proposed approach performs well on exploiting the semantic information, spatial and object context information and label dependencies.

\begin{figure}[htpb]
	\centering
	\centerline{\includegraphics[
		 width=0.9\linewidth]{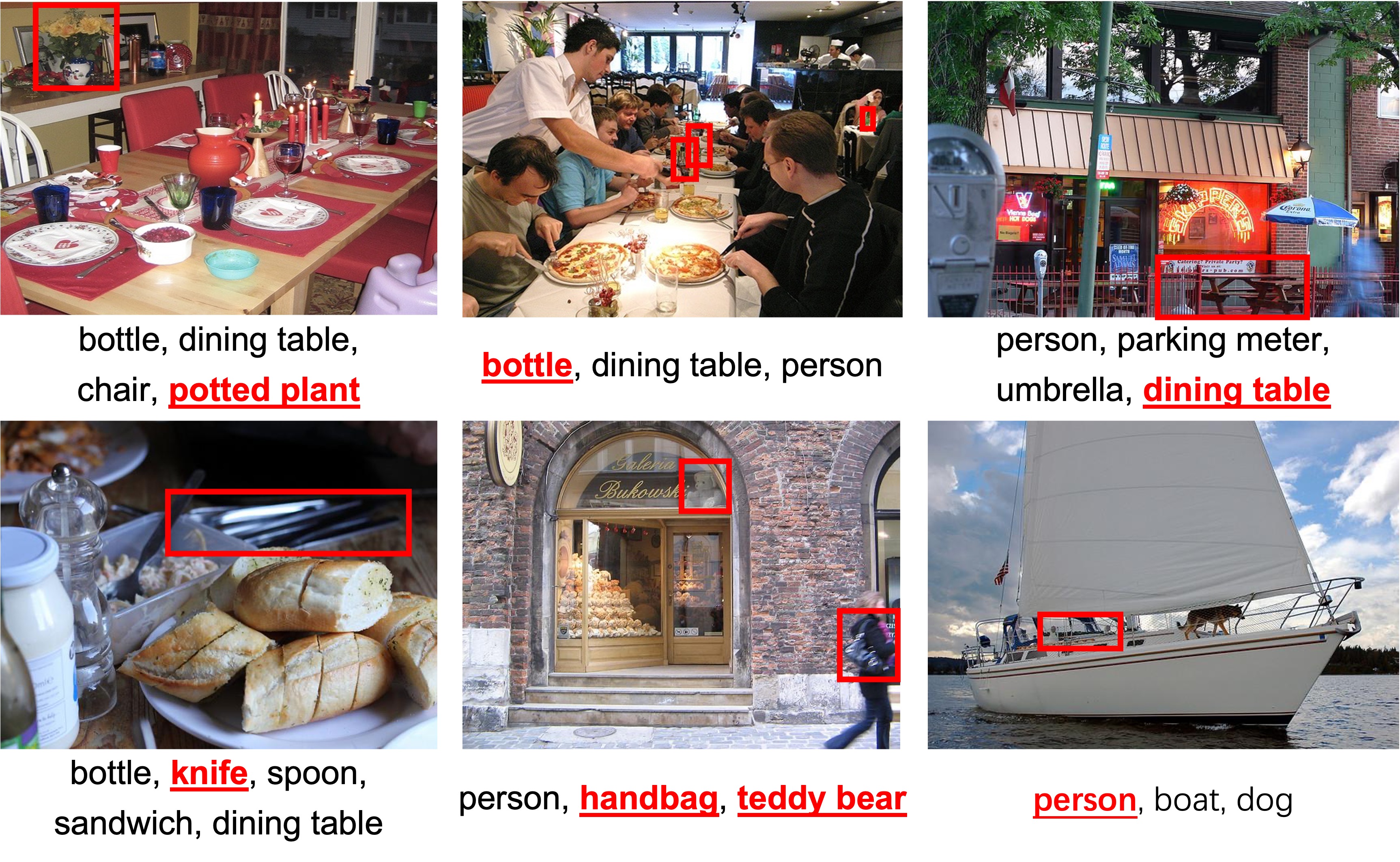}}
	\caption{Sample detection results. Small or partially occluded objects could be better detected with the background context.}
	\label{fig:visualization}
	\vspace{-0.3cm}
   \end{figure}

Since the results of the state-of-the-art models are collected from the corresponding origional paper, only eight previous solutions are selected for the performance comparison on PASCAL VOC \cite{pascal-voc-2007} dataset, and only comparing on the key evaluation metrics, mAP. The comparison results are summarized in Table \ref{tab:results_pascal}. The proposed model achieves an mAP of 95.3 $\%$, surpassing all the state-of-the-art models.

To visually demonstrate the effectiveness of the proposed model, several sample detection results for are shown in Fig. \ref{fig:visualization}. The marked labels and bounding boxes prove that the proposed method can effectively utilize the context information. Objects in red are the ones that can not be detected by previous approaches but now can be detected by the proposed model by utilizing the spatial and contextual information. From Fig. \ref{fig:visualization}, it can be seen that the proposed model plays a key role in the annotation of the easily neglected small objects such as the knife on the table and the person sailing the boat.

%


\vspace{-0.4cm}
\section{Conclusion}
\label{sec:conclusion}
\vspace{-0.1cm}

In this paper, to make use of the spatial information of objects and the contextual information around the object, a spatial-context-aware deep neural network is designed for multi-class image classification problem. The object localization and the patch expansion enable the model to leverage both the semantic and spatial information of objects. The comparisons to the state-of-the-art solutions on the MS-COCO dataset and PASCAL VOC dataset demonstrate that the proposed network significantly and consistently outperforms all compared models.

\newpage
\small
\bibliographystyle{IEEEbib}
\bibliography{strings,refs}

\end{document}